%% file: main.tex
\pdfoutput=1
\documentclass[11pt]{article}

\usepackage[final]{acl}

\usepackage{times}
\usepackage{latexsym}

\usepackage[T1]{fontenc}
\usepackage[utf8]{inputenc}

\usepackage{microtype}

\usepackage{inconsolata}

\usepackage{graphicx}

\usepackage{amsmath}
\usepackage{amsthm}
\newtheorem{lemma}{Lemma}
\usepackage{amssymb}
\usepackage{enumitem}

\usepackage{cleveref}
\usepackage{tabularray}
\usepackage{algorithm}
\usepackage{algorithmic}
\usepackage{fancyhdr}

\title{Out-of-Distribution Detection via Dropout in Polysemantic Code for Specialized LLMs}
\title{Dropout-based OOD Detection in Polysemantic Superposition Codes for Specialized LLMs}
\title{Polysemantic Dropout: Out-of-Distribution Detection for Specialized LLMs}
\title{Polysemantic Dropout: Conformal OOD Detection for Specialized LLMs}

\author{
 \textbf{Ayush Gupta\textsuperscript{1,2}},
 \textbf{Ramneet Kaur\textsuperscript{1}},
 \textbf{Anirban Roy\textsuperscript{1}},
 \textbf{Adam D. Cobb\textsuperscript{1}},
\\
 \textbf{Rama Chellappa\textsuperscript{2}},
 \textbf{Susmit Jha\textsuperscript{1}},
\\
\\
 \textsuperscript{1}SRI,
 \textsuperscript{2}Johns Hopkins University,
}

\definecolor{ForestGreen}{RGB}{34,139,34}

\begin{document}
\thispagestyle{fancy}
\maketitle
\begin{abstract}
We propose a novel inference-time out-of-domain (OOD) detection algorithm for specialized large language models (LLMs). Despite achieving state-of-the-art performance on in-domain tasks through fine-tuning, specialized LLMs remain vulnerable to incorrect or unreliable outputs when presented with OOD inputs, posing risks in critical applications. Our method leverages the Inductive Conformal Anomaly Detection (ICAD) framework, using a new non-conformity measure based on the model's dropout tolerance. Motivated by recent findings on polysemanticity and redundancy in LLMs, we hypothesize that in-domain inputs exhibit higher dropout tolerance than OOD inputs. We aggregate dropout tolerance across multiple layers via a valid ensemble approach, improving detection while maintaining theoretical false alarm bounds from ICAD. Experiments with medical-specialized LLMs show that our approach detects OOD inputs better than baseline methods, with AUROC improvements of $2\%$ to $37\%$ when treating OOD datapoints as positives and in-domain test datapoints as negatives.
\end{abstract}

\input{1_intro}
\input{3_background}

\input{4_tech}

\input{5_exp}

\input{2_rel}
\input{6_conc}

% \section{Appendices}

% Use \verb|\appendix| before any appendix section to switch the section numbering over to letters. See Appendix~\ref{sec:appendix} for an example.

\newpage
\section*{Limitations}
As discussed in the ablation study, the proposed non-conformity measure, and hence the $p$-value is undefined if the original response to the query does not change even after dropping the maximum number of neurons for the proposed algorithm. Although this situation was observed only rarely in our case studies and can be mitigated by increasing the number of layers used in the ensemble approach, it may still occur and cause the algorithm to incorrectly classify an OOD input as in-distribution. Increasing the number of layers further improves the OOD detection performance but it also means higher computational cost at inference time. 

\section*{Acknowledgments}
This material is based upon work supported by the United States Air Force and DARPA under Contract No. FA8750-23-C-0519 and HR0011-24-9-0424, and the U.S. Army Research Laboratory under Cooperative Research Agreement W911NF-17-2-0196 and Defense Logistics Agency
(DLA) and the Advanced Research Projects Agency for Health (ARPA-H) under Contract Number
SP4701-23-C-0073. Any opinions, findings
and conclusions or recommendations expressed in this material are those of the author(s) and do not necessarily reflect the views of the United States Air Force, DARPA, the U.S. Army Research Laboratory, ARPA-H or the United States Government.

\bibliography{custom}

\appendix

\input{appendix_tex}

\end{document}

%% file: 1_intro.tex
\section{Introduction}
\label{sec:intro}
% Link to the submission website - \url{https://2025.emnlp.org/calls/main_conference_papers/}

\begin{figure}
    \centering
    \includegraphics[width=\linewidth]{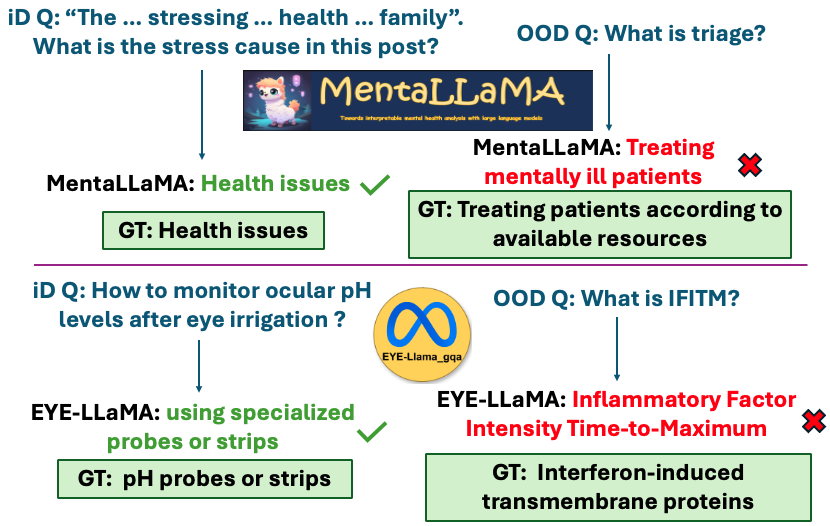}
    \caption{LLMs specialized in Medical Domain work well on in-domain queries, but are prone to make mistakes on out-of-domain (OOD) queries: MentaLLaMA associates its responses to OOD queries with mental health, and EYE-LLaMA hallucinates on those. Here `GT' stands for the Ground Truth Answer.
    }
    \label{fig:motivation}
    \vspace{-0.5cm}
\end{figure}

LLMs’ ability to generate coherent, contextually relevant and often human-level language has led to their rapid adoption in industry and research, powering applications such as recommendation systems~\citep{llm_recommendation}, legal analysis~\citep{legal_bert}, literature review~\citep{llm_lit_review}, drug discovery~\citep{llm_drug}, and clinical decision support~\citep{llms_medicine}. When these models are fine-tuned for specialized tasks, they achieve state-of-the-art performance by leveraging the domain-specific knowledge~\citep{fine_tuned_sota_llms}. However, these fine-tuned LLMs remain susceptible to errors when confronted with data that falls outside the scope of their domain. Fig.~\ref{fig:motivation} shows incorrect responses by two medical LLMs when prompted for out-of-domain (OOD) questions. According to our analysis, MentaLLaMA~\citep{mentalllama}: an LLM specialized in mental health analysis, associates most of its responses on OOD inputs with mental health. On the other hand, EYE-LLaMA~\citep{eyellama}: an LLM specialized in ophthalmology, mostly hallucinates on these OOD queries. 

\begin{figure*}
    \centering
    \includegraphics[width=\linewidth]{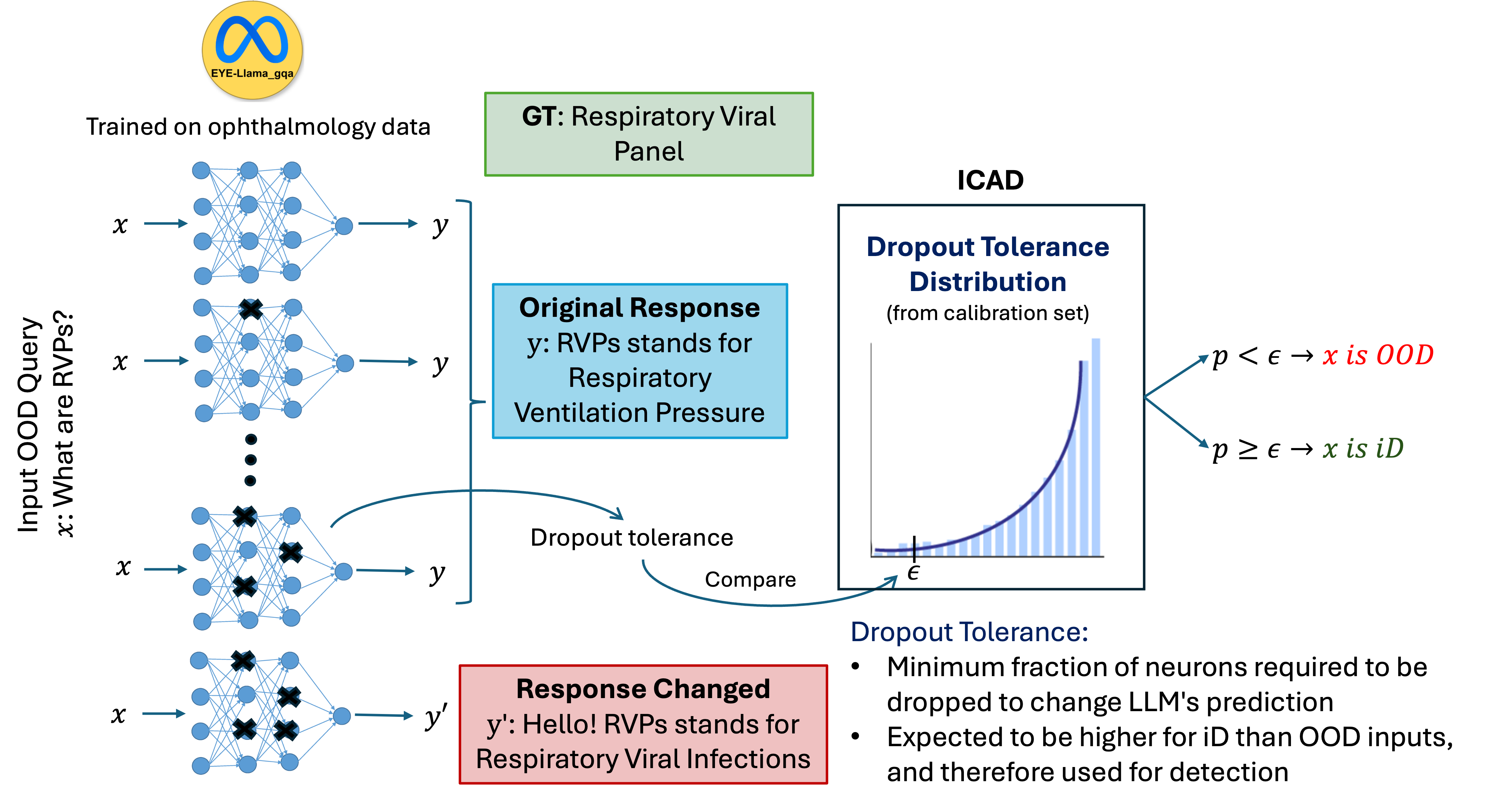}
    \caption{An overview of the proposed approach for OOD detection for specialized LLMs: We propose to leverage \textit{dropout tolerance} for non-conformity measure in the Inductive Conformal Anomaly Detection (ICAD) framework for detection with bounded false alarm rate.}
    \label{fig:framework}
    \vspace{-0.5cm}
\end{figure*}

In this paper, we address the challenge of detecting OOD inputs for specialized LLMs, aiming to enhance their reliability and safety in real-world applications. We leverage the Inductive Conformal Anomaly Detection (ICAD) framework~\citep{icad} for OOD detection with bounded false alarms. Central to this framework is the non-conformity measure (NCM), a real-valued function that quantifies the non-conformity of an input to the training distribution.  The success of ICAD depends on the choice of NCM: a good measure that can distinguish between in and out-of-domain inputs. We propose to utilize the dropout tolerance of these specialized LLMs as the NCM in the ICAD framework for OOD detection. We define the dropout tolerance of an LLM on an input query $x$ as the minimum fraction of neurons required to be dropped from a layer of the model to change its original prediction on $x$. Fig.~\ref{fig:framework} gives an overview of the proposed approach.

LLMs are expected to be robust to perturbations such as dropout due to redundancy or distribution of concepts across neurons. Polysemantic code learned by these networks favors redundancy~\citep{poly_dropout}, and we hypothesize this redundancy to be higher for in-domain than OOD inputs for LLMs specialized in a particular domain. Polysemanticity~\citep{sae_interpretability} is a widely investigated topic in the mechanistic interpretability research community. It refers to the phenomenon where neurons activate on multiple concepts to maximize the model’s capacity, thereby making these models challenging to interpret. The contributions of this paper can be summarized as follows:\\
\textbf{1. Novel NCM:} We propose a novel NCM based on dropout tolerance of LLMs in the ICAD framework for detection of out-of-domain or out-of-distribution detection for specialized LLMs\footnote{We use `OOD' for out-of-domain or out-of-distribution  in the paper.}.\\
\textbf{2. Ensemble Approach:} Instead of relying on a single layer for the dropout, we propose an ensemble approach where detections from different layers can be combined by a valid merging function while preserving the false alarm rate guarantees of the ICAD framework. \\
% \AG{Should we stress that our method is inference-time, and doesn't require any training? Maybe we should take advantage of this throughout the introduction}\\ 
\textbf{3. OOD Detection Algorithm:} We propose an inference-time OOD detection algorithm based on the ensemble approach with the proposed NCM.
%and changing of the response after dropout. \AG{Check point 3} \\  
% with the specifics of on dropout for the proposed NCM, checking for the change in response \\
\textbf{4. Experimental Evaluation:} We perform extensive experiments on LLMs specialized in the medical domain: MentaLLaMA and EYE-LLaMA, on multiple OOD datasets. We compare AUROC and ROC results on OOD detection with three baselines, and empirically evaluate false alarm guarantees along with several ablation studies for the proposed algorithm.

% In comparison with our baselines, we achieve state-of-the-art results with $2\%$ to $37\%$ performance gains in AUROC. 

%% file: 3_background.tex
\section{Background}
\label{sec:back}
\subsection{Polysemanticity and Dropout}
\textit{Polysemanticity} refers to the phenomenon where individual   neurons within LLMs activate on multiple, often disparate concepts or features. The hypothesized cause of polysemanticity is \textit{superposition}, where these models encode far more features than neurons. This is accomplished by distributing features to an over-complete set of directions in the activation space rather than to individual neurons to maximize the model's capacity, making it difficult to interpret LLMs~\citep{sae_interpretability}.

~\citet{poly_dropout} connect information theory with polysemanticity  where polysemantic code learned by LLMs is not only efficient but also favors \textit{redundancy} for robustness (Fig. 2 of their paper). Redundancy refers to the distribution of features across neurons, therefore, discouraging monosemantic code
\footnote{Monosemanticity refers to the mapping of a concept or feature to a single neuron.} 
and making the model robust to noise or perturbations such as dropout. Dropout is a technique that drops a fraction of neurons from the neural network while making predictions on an input. It was introduced as a regularization technique to avoid overfitting during the training phase where a neuron would be dropped with the probability $p$ (dropout rate) at each training iteration, and weights of all neurons would be scaled down by the dropout rate during inference~\citep{dropout}. The idea is to prevent the network from becoming too dependent on certain nodes, or in other words, promote redundancy of features among nodes for generalizability and robustness.

% Dropout prevents the network from becoming too dependent on certain nodes and encourages it to learn more generalized features, which helps it perform better on new data.

% Dropout a technique that randomly disables (or “drops”) a fraction of neurons during each training iteration. This prevents the network from becoming too dependent on certain nodes and encourages it to learn more generalized features, which helps it perform better on new data.

% From the paper: Thus, when a neural network is exposed to moderate noise, we expect the code to become more redundant, distributing the representation of inputs across multiple activation patterns. While this distribution of representations across neurons discourages a monosemantic code (Fig. 2a), it is unclear whether it might facilitate the holistic interpretation of activations across a given layer.

% 1st attempt: ~\citet{poly_dropout} connect information theory with polysemanticity  
% where polysemantic code learned by the network is not only efficient but also  favorable for robustness via \textit{redundancy} (Fig. 2 of their paper). Redundancy is a property of the robust neural network, where features are spread out across neurons such that if we drop a few of those, the network will still be able to reconstruct features from the surviving neurons. 

% \subsection{Dropout}
% % \url{https://medium.com/codex/understanding-dropout-in-deep-neural-networks-95e7d1b11c58}, \url{https://medium.com/@piyushkashyap045/understanding-dropout-in-deep-learning-a-guide-to-reducing-overfitting-26cbb68d5575}

\subsection{Conformal Prediction}
Conformal prediction~\citep{cp} is a statistical framework used to assess the degree to which a new input conforms to the training distribution. Central to this framework is the non-conformity measure (NCM), a real-valued function that quantifies the non-conformity of an input $(x)$ with respect to the training distribution by assigning it a non-conformity score $\alpha_x$. Given a training dataset $X = \{x_1, x_2, \ldots, x_l\}$, the NCM evaluates how atypical an input is with respect to $X$, with larger scores indicating a higher degree of non-conformity. A variety of NCMs have been proposed in literature, employing methods such as $k$-nearest neighbors~\citep{icp}, support vector machines~\citep{icp}, random forests~\citep{ncm-rf}, variational autoencoders~\citep{vanderbilt}, memory prototypes~\citep{yang2024memory}, transformation equivariance~\citep{idecode, kaur2024tcps}. 

Conformal anomaly detection (CAD)~\citep{cad} makes use of the NCM to flag inputs anomalous to the training distribution from its $p$-value. 
The $p$-value of an input $x$ is computed by comparing its non-conformity score $\alpha_x$ to these scores of the training samples:
\begin{equation*} 
p\text{-}value = \frac{|{i=1,...,l: \alpha_{x} \leq \alpha_i}|+1}{l+1}, 
\end{equation*}
where $\alpha_i$ denotes the non-conformity scores of the training data, calculated using the NCM defined on the new dataset constructed from the training dataset of size $l$ and the new input $x$. If $x$ follows the same distribution as the training data, its score is expected to lie within the range of these scores for the training data, resulting in a higher $p$-value. Conversely, $x$ is flagged as anomalous if its $p$-value falls below a chosen detection threshold $\epsilon \in (0,1)$.

% A drawback of this approach is the computational inefficiency associated with recalculating non-conformity scores for the entire set each time a new input is encountered, especially in scenarios where NCM is computationally expensive. 
In scenarios where the NCM is computationally expensive, recalculating scores for the entire dataset upon the arrival of each new input imposes a substantial computational overhead. To address this limitation of CAD, inductive conformal anomaly detection (ICAD) was introduced~\citep{icad}. In ICAD, the training set is partitioned into a proper training set $X_{\mathrm{tr}} = \{x_1, \ldots, x_m\}$ and a calibration set $X_{\mathrm{cal}} = \{x_{m+1}, \ldots, x_l\}$. NCM is defined on $X_{\mathrm{tr}}$, and the $p$-value for a new input $x$ is calculated by comparing its non-conformity score $\alpha_x$ to those computed for the calibration set:
\begin{equation} p\text{-}value = \frac{|{ i=m+1,...,l : \ \alpha_{x} \leq \alpha_{i}}|+1}{l-m+1}. \label{eq:p_value} \end{equation}
Scores for the calibration set are precomputed offline and used during inference, improving the computational efficiency of the CAD framework. Again, an input is considered anomalous if its $p$-value is less than the detection threshold $\epsilon$. 

\begin{lemma}~\citep{cp}
If the test input $x$ and the calibration points are independent and identically distributed (i.i.d.), the $p$-value computed in~\eqref{eq:p_value} is uniformly distributed in $(0, 1)$. Therefore, the probability of a false alarm (i.e., erroneously labeling a non-anomalous input as anomalous) is upper bounded by the detection threshold $\epsilon$. 
\label{lemma:uniform_p_values} 
\end{lemma}

The success of ICAD depends on the choice of NCM used in the framework. We propose to leverage dropout tolerance, i.e. the minimum fraction of neurons that must be deactivated to alter LLMs's prediction. Our hypothesis is that these polysemantic models exhibit greater robustness (i.e., can tolerate higher levels of dropout) on in-distribution (iD) inputs compared to OOD inputs.

% \textcolor{blue}{The success of ICAD depends on the choice of the NCM. We propose dropout...motivated from \url{https://arxiv.org/pdf/2401.17975} where polysemantic neurons can tolerate dropout to some extent!}

\subsection{Combining hypothesis}
\label{sec:combining-hypothesis}
Same hypothesis of ``an input drawn from the training distribution'' can be tested using multiple conformal anomaly detectors, and merging their results with an ensemble approach can lead to better performance than individual detectors. Multiple $p$-values $(p_1,\ldots,p_K)$ of an input from~\eqref{eq:p_value} by $K$ conformal anomaly detectors can be combined using the following averaging function $M_{r,K}$:
% ~\citet{vovk2020combining_p_values} propose a number of merging functions that can be used to combine $p$-values from different detectors testing the same hypothesis without violating detection guarantees from ICAD. 
\begin{align}
M_{r,K}(p_1,\ldots,p_K) &= \left ( \frac{p_1^r+\ldots+p_K^r}{K} \right )^{1/r}, \notag
\end{align}

% &\text{where } r \in \mathbb{R} \setminus \{0\}
with the special cases of $r=0$, $\infty$, and $-\infty$ defined as follows:
\begin{align}
    M_{0,K}(p_1,\ldots,p_K) &= \exp \left ( \frac{\ln p_1+\ldots+\ln p_K}{K} \right ) \notag \\             &=\left ( \prod_{k=1}^K p_k \right )^{1/K},
    \label{eq:geom_avg_func}
\end{align}

\begin{equation*}
    M_{\infty,K}(p_1,\ldots,p_K) = \max(p_1,\ldots,p_K).
\end{equation*}
\begin{equation}
    M_{-\infty,K}(p_1,\ldots,p_K) = \min(p_1,\ldots,p_K),
    \label{eq:bon_avg_func}
\end{equation}
~\citet{vovk2020combining_p_values} make use of $M_{r,K}$ for defining \textit{valid} merging functions on $K$ $p$-values: 
\begin{equation}
    a_{r,K} M_{r,K}(p_1,\ldots,p_K), \ r \in [-\infty,\infty], K \geq 2.
    \label{eq:merging_func}
\end{equation}
Here $a_{r, K}$ is a constant required for making the merging function valid, i.e. preserving the false alarm rate guarantees of ICAD (Lemma~\ref{lem:vovk_mf}).
% The value obtained by applying a \textit{valid merging function} on the $K$ uniformly distributed $p$-values is also uniformly distributed. 
We consider the following four merging functions for combining $K$ $p$-values computed from dropout in $K$ layers of the LLM. These functions vary in the value of $r$~\citep{vovk2020combining_p_values}:
\begin{enumerate}[leftmargin=*]
    \item \textbf{Harmonic Mean (HM)}: With $r=-1$, $a_{-1,K} = (\ln K)$, and $M_{-1,K}(p_1,\ldots,p_K) = \left ( \frac{p_1^{-1}+\ldots+p_K^{-1}}{K} \right )^{-1}.$ 
    \item \textbf{Arithmetic Mean (AM)}: With $r=1$, $a_{1,K} = (1+r)^{1/r} = 2$, and $M_{1,K}(p_1,\ldots,p_K) = \left ( \frac{p_1+\ldots+p_K}{K} \right ).$
    \item \textbf{Geometric Mean (GM)}: With $r=0$, $a_{0,K} = e$, and $M_{0,K}(p_1,\ldots,p_K)$ is as defined in ~\eqref{eq:geom_avg_func}.
    \item \textbf{Bonferroni Method (BM)}: With $r=-\infty$, $a_{-\infty,K} = K$, and $M_{-\infty,K}(p_1,\ldots,p_K)$ is as defined in ~\eqref{eq:bon_avg_func}.
\end{enumerate}

% \begin{lemma}[\citeauthor{vovk2020combining_p_values}, \citeyear{vovk2020combining_p_values}]
%     For all the $K$ $p$-values uniformly distributed in $[0, 1]$, the value obtained by applying merging function from~\eqref{eq:merging_func} on these $K$ $p$-values is a valid $p$-value, i.e., the combined $p$-value is also uniformly distributed in $[0, 1]$, and therefore $Pr\left( 
%     \text{combined } p\text{-value} < \epsilon \right) \le \epsilon, \forall \epsilon \in (0,1)$. Validity of the combined $p$-value holds true without assuming any dependence structure among the $K$ $p$-values.
%     \label{lem:vovk_mf}
% \end{lemma}

\begin{lemma}[\citeauthor{vovk2020combining_p_values}, \citeyear{vovk2020combining_p_values}]
If $p_1, \ldots, p_K$ are uniformly distributed in $[0,1]$, then the value obtained by applying a merging function from~\eqref{eq:merging_func} is a valid $p$-value—meaning it is uniformly distributed in $[0,1]$. Thus, for any $\epsilon \in (0,1)$, we have $Pr(\text{merged $p$-value} < \epsilon) \le \epsilon$. This validity holds without any assumptions about the dependence among the $K$ $p$-values.
\label{lem:vovk_mf}
\end{lemma}

%% file: 4_tech.tex
\section{OOD Detection for Specialized LLMs}
\label{sec:tech}
\textbf{Proposed NCM:} 
NCM assigns a score which measures non-conformity of an input with respect to the training distribution. So, it is expected to be higher for OOD and lower for in-distribution (iD) inputs. 
We, therefore, propose to use $1-$\textit{dropout tolerance} of an LLM's layer as the NCM. 
Here, dropout tolerance for a layer $L$ is defined as the minimum fraction of neurons in $L$ that must be dropped to change the LLM’s original prediction.
% \begin{equation*}
%    \frac{\text{\#neurons dropped in $L$ to change the prediction}}{\text{\#neurons in $L$}}
% \end{equation*}
By original prediction, we mean the prediction made by the model without any dropout. 

The intuition behind this score is that an LLM specialized in a particular domain is expected to have higher dropout tolerance for iD inputs than OOD inputs, resulting in higher non-conformity score for OOD than iD inputs. We propose to use this score in the ICAD framework~\eqref{eq:p_value} for OOD detection with bounded false alarms.

\textbf{Ensemble Approach:} Instead of relying on a single layer for OOD detection, we propose an ensemble approach where $p$-values~\eqref{eq:p_value} from different layers can be combined by valid merging functions~\citep{vovk2020combining_p_values}. We consider the four merging functions defined in Section~\ref{sec:combining-hypothesis} on the $K$ $p$-values computed from $K$ layers by using the proposed NCM for each layer; thereby preserving the theoretical guarantees from the ICAD framework. 

\textbf{Proposed Algorithm for OOD Detection:} With different ways of choosing the neurons to be dropped, comparing semantics of the original response with the one after dropout etc., there can be different ways of implementing an OOD detection algorithm with the merged $p$-value from the proposed NCM. We describe the specifics of different steps in the proposed Algorithm~\ref{alg:ood_detection} as follows.

\textit{1. Selection of neurons to be dropped:} Based on the activations of a layer while generating the last token on an input query $x$, we construct the list $L$ of the $m$ most activated neurons in the layer
% , based on their activations when generating the last token of the input query $x$
\footnote{$L$ contains $m$ maximally activated neurons stored in ascending order of activation.}. We choose the last token as it captures context for the entire response.

\textit{2. Iterative dropout:} 
% We ask the same query to the LLM multiple times, each time dropping $n$ more neurons sequentially from the list $L$. 
% In each iteration, we check if the response of the LLM is semantically similar to the original pre-dropout response.
We query the LLM for $x$ multiple times, each time dropping $n$ additional neurons from $L$, and checking if the generated response after the dropout is semantically similar to the pre-dropout response from the model.

\textit{3. Checking for change in the response:} After each dropout iteration, we prompt GPT-4o~\citep{gpt-4} to evaluate whether the generated response is semantically similar to the pre-dropout response. 
% If the responses are similar, then we increase the number of neurons to be dropped in the next iteration.
If the response is the same, we continue to the next iteration. Otherwise, we compute the \textit{dropout tolerance} on $x$ as the fraction of neurons dropped to change the pre-dropout response on $x$. 

\textit{4. OOD Detection:} We compute the non-conformity score $\alpha_x$ as $1-$\textit{dropout tolerance} of the LLM on $x$. We compare $\alpha_x$ to these scores of all queries in the calibration set (computed offline), to obtain the $p$-value for $x$. 
We do this for ($K$) layers in the model for getting $K$ $p$-values, calculate the merged $p$-value $p_{merged}$ from~\eqref{eq:merging_func}, and compare it with the detection threshold $\epsilon$. 
$x$ is detected as OOD if $p_{merged} < \epsilon$, and iD otherwise. 

\begin{algorithm}[t]
\caption{OOD Detection for Specialized LLMs}
\label{alg:ood_detection}
\begin{algorithmic}[1]
    \STATE \textbf{Input:} Specialized LLM $M$, Input query $x$, Maximum number $m$ of neurons to be dropped from a layer, $K$ layers selected for dropout for the ensemble approach, Merging function $M_{r,K}$, $K$ sets of calibration set alphas $\{\alpha_j^{k}: 1 \leq k \leq K, m+1\le j \le l\}$, Evaluation LLM $E$, detection threshold $\epsilon$
    \STATE {\bfseries Output:} ``$1$'' if $x$ is detected as OOD; ``$0$'' otherwise 
    \STATE $y_{orig} = M(x)$ \COMMENT{Original response}
    \STATE Initialize $k = 1$ \COMMENT{For iteration over layers}
    \WHILE{$k \leq K$} 
        \STATE $L = $ list of $m$ maximally activated neurons in layer $k$ on last token of $y_{orig}$
        \STATE Initialize $i = n$ \COMMENT{For iterative dropout}
        \WHILE{$i<m$}
            \STATE Drop the first $i$ neurons from $L$
            \STATE $y_{dropout} = M_{dropout}(x)$ %\COMMENT{Response after dropping $i$ neurons}
            \IF{$E(y_{current}, y_{dropout}) == \text{"different"}$}
                \STATE Goto line $16$
            \ENDIF
            \STATE $i = i+n$
        \ENDWHILE
        %\STATE Let $k_{stop} = k$ 
        \STATE $\alpha_{x}^{k} = 1 - \frac{i}{\text{total \#neurons in }k}$
        % \STATE Compute Non-conformity Scores $\alpha_i = \alpha(x_i)$ for all $x_i \in C$
        \STATE $p_k = \frac{|{ j=m+1,...,l : \ \alpha_{x}^{k} \leq \alpha_{j}^{k}}|+1}{l-m+1}$
    \ENDWHILE
    \STATE $p_{merged} = M_{r,K}(p_1,\ldots, p_K)$
    \STATE \textbf{If} $p_{merged} < \epsilon$ \textbf{then} \textbf{return} $1$ \textbf{else} \textbf{return} $0$ 
    %     \STATE \textbf{Output:} OOD
    % \ELSE
    %     \STATE \textbf{Output:} in-distribution
    % \ENDIF
\end{algorithmic}
\end{algorithm}
%\vspace{-3mm}

%% file: 5_exp.tex
\section{Experiments}
\label{sec:exp}
\subsection{Specialized LLMs and iD datasets} We consider two LLMs specialized in particular domains of healthcare: EYE-LLaMA and MentaLLaMA. Details of these models and their in-distribution (iD) datasets are as follows.
\paragraph{EYE-LLaMA}~\citep{eyellama} is a specialized LLM developed to enhance natural language understanding and QA capabilities within the field of ophthalmology. Built upon LLaMA 2~\citep{llama2}, EYE-LLaMA addresses the unique linguistic and informational needs of ophthalmic practitioners, researchers, and educators. EYE-LLaMA was trained in two phases: pretraining on 766K ophthalmology documents and fine-tuning on the EyeQA dataset.
% We use the EYE-LLaMA\_gqa variant in our experiments.
% \textbf{Pre-training Phase}: The base model, referred to as EYE-LLaMA\_p, was pre-trained on an extensive corpus of approximately 766,000 ophthalmology-related documents. This corpus includes abstracts from PubMed, textbooks, articles from EyeWiki and relevant entries from Wikipedia. Pre-training phase aims to provide the model with a comprehensive understanding of ophthalmic terminology, concepts, and literature.
% \textbf{Fine-tuning Phase}: Following pre-training, the model underwent fine-tuning to specialize in question-answering tasks. Two versions were developed, EYE-LLaMA\_qa and EYE-LLaMA\_gqa, with the latter being generated by further fine-tuning using GPT-generated samples. EYE-LLaMA\_gqa is regarded as the better model and we use this version in our experiments.
% \AG{Describe EyeQA dataset - subjective and MCQ queries.}

\textbf{EyeQA (iD Dataset for EYE-LLaMA)}~\cite{eyellama} amalgamates approximately 744,000 unsupervised text samples sourced from PubMed abstracts, 22,000 samples from nearly 570 textbooks, and articles from EyeWiki and Wikipedia's ophthalmology category as the pretraining dataset.
For supervised fine-tuning, the dataset includes around 18,000 QA pairs from medical datasets, 1,500 QA pairs from medical forums and is further enriched by 15,000 QA pairs generated by GPT-3.5.
This dataset contains a mix of multiple-choice and descriptive queries and was used to fine-tune EYE-LLaMA - to make it specialized for the ophthalmology domain.
\paragraph{MentaLLaMA}~\citep{mentalllama} is an open-source instruction-following LLM developed for interpretable mental health analysis on social media data. 
Again built upon the LLaMA 2 architecture, MentaLLaMA is designed to perform mental health classification tasks while providing human-readable explanations for its predictions. 
% MentaLLaMA can also provide explanations about the answers it provides. 
Multiple MentaLLaMA versions -- 7B, 13B and 33B -- are available.
We utilize the MentaLLaMA-chat-7B variant in this work. 
This model's training dataset, IMHI, is described below.

\textbf{IMHI (iD Dataset for MentaLLaMA)}
\cite{mentalllama} curate a dataset called the Interpretable Mental Health Instruction (IMHI) dataset to train MentaLLaMA. IMHI is a multi-task, and multi-source corpus designed to facilitate instruction-tuning of LLMs for interpretable mental health analysis.  Comprising approximately 105,000 instruction-response pairs, the dataset has distinct mental health tasks—including depression detection, stress cause identification, and wellness classification—sourced from platforms such as Reddit and Twitter. The dataset contains long descriptive questions, with answers and detailed reasoning behind the answers generated by ChatGPT~\cite{mentalllama}.

\subsection{OOD datasets}
% We evaluate our approach on both in-domain and OOD queries for a model. The in-domain queries are sampled from the respective training datasets of the model and the OOD queries from other QA datasets.
We use COVID-QA and MedMCQA as the OOD datasets in our work and evaluate our approach on both LLMs using both of these OOD datasets. 

\textbf{COVID-QA}~\citep{covidqa} is a specialized dataset comprising 2,019 QA pairs annotated by 15 biomedical experts. 
These annotations are derived from 147 scientific articles focusing on COVID-19-related content. 
Each entry includes a question, a contextual passage from the source article, and a corresponding answer, formatted in the SQuAD style~\cite{squad}.
The dataset features contexts averaging around 6,000 tokens and answers averaging 14 words.
%This design aims to enhance the development of domain-specific QA models for biomedical literature, particularly in the context of COVID-19 research.

\textbf{MedMCQA}~\cite{medmcqa} is a large-scale MCQ dataset, comprising over 194,000 high-quality questions sourced from AIIMS and NEET PG exams across 21 medical subjects such as Anatomy, Pathology, Pharmacology, and Surgery.
Each question includes four answer options and an explanation. 
% The dataset is designed to evaluate and enhance the reasoning abilities of models across a variety of medical topics, emulating the rigor of real-world medical exams.
Notably, this dataset contains ophthalmology questions, which we filter when using it as OOD for EYE-LLaMA.

We also report results with EYE-QA and IMHI datasets as OOD datasets for MentaLLaMA and EYE-LLaMA respectively, in the Appendix. 

\subsection{Metrics}
With OOD inputs as positives and iD test datapoints as negatives, we compare the Receiver Operating Characteristic (ROC) curves and the corresponding Area Under the Curve (AUROC) for both LLMs on both OOD datasets against the baselines. We also report the false alarm rate guarantee curves by Algorithm~\ref{alg:ood_detection} for both models and OOD datasets.

\subsection{Baselines}
We compare our method with three baselines involving the iterative dropout procedure. These baselines are described below.
%\AG{Can we compare with any existing method for OOD detection? If not, say that we are the first work in this domain.}

% \textbf{Hard thresholding on $\alpha$.}
% As mentioned in \Cref{sec:tech}, for a given query, we obtain a corresponding non-conformity score $\alpha$ after performing iterative dropout. 
% In this baseline, we directly compare $\alpha$ with a hard threshold to classify the sample as in-distribution or out-of-distribution. 
% This differs from our method because of the lack of p-value calculation and the lack of a calibration set.
\textbf{1. Base Score Method:} Score $\alpha_x$ from the proposed NCM can also be directly used for OOD detection without the ICAD framework (or computing $p$-value from $\alpha_x$). We use these non-conformity scores for detection and refer to this baseline as the \textit{base score} method. This method, however, does not provide any guarantees on false alarms.

% \textbf{Individual layers.}
% We also compare our method with variants without the aggregation step. Here, a query's $p$-value is calculated from a single layer.

\textbf{2. Single $p$-value Method:} In this baseline, we use the traditional ICAD approach with just one $p$-value. This $p$-value is calculated from the proposed NCM with dropout in a single layer.

\textbf{3. Ensemble Approach with Majority Voting:}
Here, we use majority voting instead of a valid merging function on the individual $p$-values from different layers. Specifically, we run the single $p$-value method on different layers and use majority voting on those detections. 
This baseline also does not provide any false alarm rate guarantees. 

% In this baseline, instead of obtaining an aggregated p-value based on the $\alpha_x$ for each layer, we get a final prediction through a voting mechanism between the three layers.   Specifically, we let each layer make a prediction (iD / OOD) separately and vote on a query. The majority vote is the prediction.
% It should be noted that this method is not a valid p-value merging function and hence does not provide theoretical guarantees on false alarms. 

% Baseline - just use NCS [hard threshold, no cal set, no p-value setup] as it is for detection, one single layer 
% individual layers p-value as baselines
% voting mechanism for combining without guarantees on same NCS.

% \subsection{Metrics}
% Explain how to plot ROC. (Maybe not needed?) 
% \AG{Decide after you write the methodology section.}

\subsection{Results}
\label{main_results}
\begin{figure*}[!t]
    \centering
    \includegraphics[width=\linewidth]{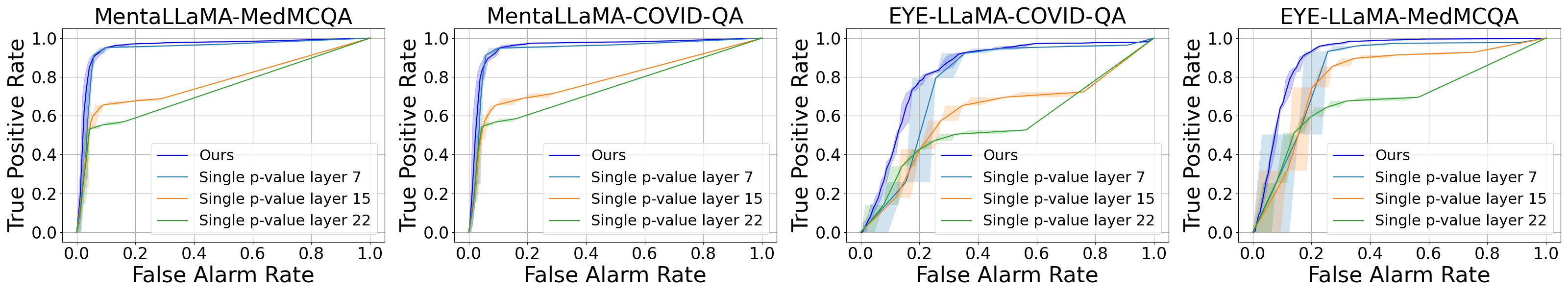}
    \vspace{-7mm}
    \caption{Comparison of ROC curves for OOD Detection with the single $p$-value baselines across layers. 
    %The ROC curves are shown in blue, and the false positive rate - epsilon curves are shown in green. A reference x=y line is shown in red. 
    % Our method performs better than the baselines on the OOD detection task.
    %\AG{Increase tick size, correct mentallama spelling}
    %\AG{False alarm - label it correctly}
    }
    \label{fig:main-plots}
\end{figure*}

With the number of layers $K = 3$ -- specifically, layers $7$, $15$, and $22$ --  we compare AUROC results with baselines for both the models in \Cref{tab:main-results}.
% The rationale for selecting these layers, along with additional implementation details, is provided in the Appendix. 
We choose layers $7$, $15$, and $22$ as each of these layers lies at a different stage of inference: starting, middle, and towards the end, and hence the model has a different understanding of the query at each of these individual layers~\citep{stages_of_inference}. 
Our approach consistently outperforms all baselines across all test cases. We also compare ROC curves with the single $p$-value method in \Cref{fig:main-plots}. Again, the proposed approach achieves the best results with comparable performance from the single $p$-value method with Layer $7$.

\begin{figure*}[!t]
    \centering
    \includegraphics[width=\linewidth]{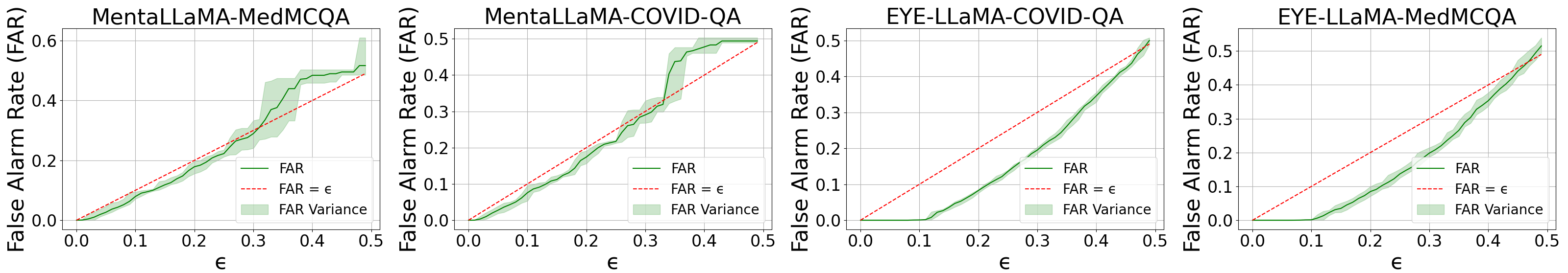}
    \vspace{-7mm}
    \caption{False Alarm Rate Guarantees Plots: False alarm is upper bounded by $\epsilon$ on average in most test cases. 
    % False alarm curves for our method, over multiple folds. The Y-axis represents the false positive rate, and the X-axis represents different levels of $\epsilon$.
    % A reference x=y line is shown in red. 
    % Our method is able to provide reasonable false positive rate guarantees by varying the user defined parameter $\epsilon$.
    % %\AG{Move this to analysis section, under a 'guarantees' subsection? This is not part of main results}
    % \AG{Update caption}
    }
    \label{fig:guarantee-curves}
    \vspace{-0.25cm}
\end{figure*}

Plots on the bounded false alarm rate guarantees by Algorithm~\ref{alg:ood_detection} are shown in \Cref{fig:guarantee-curves} with the range of detection threshold $\epsilon = \{0, 0.05, 0.1, \ldots, 0.5\}$. These plots show that the false alarm rate is upper bounded by the detection threshold for all values of $\epsilon$ for EYE-LLaMA. For MentaLLaMA, the guarantees are satisfied for most of the cases except for the higher range of $\epsilon$. This can be attributed to the statistical insignificance of the empirical calibration data representing the training distribution.

All reported results are averaged over five runs with random $80\%-20\%$ splits of the iD test and calibration sets, and we observe low standard deviation in all cases, typically as low as $0.001$.
% \AG{Std dev is 0.001. Variance will be 1e-6}
% Our method is able to provide reasonable false alarm guarantees compared to the confidence threshold $\epsilon$.
% We observe that our method outperforms all baselines across all test cases.
\begin{table*}[t]
\centering
\resizebox{0.8\linewidth}{!}{%
\begin{tabular}{c|cc|cc}
\hline
\textbf{Model} & \multicolumn{2}{c|}{\textbf{EYE-LLaMA}} & \multicolumn{2}{c}{\textbf{MentaLLaMA}} \\
\hline
\textbf{OOD Dataset} & \textbf{CovidQA} & \textbf{MedMCQA} & \textbf{CovidQA} & \textbf{MedMCQA} \\ 
\hline
Base Score Method with Layer 7 & 0.53 & 0.54 & 0.73 & 0.72 \\
Base Score Method with Layer 15 & 0.48 & 0.58 & 0.71 & 0.69 \\
Base Score Method with Layer 22 & 0.48 & 0.57 & 0.70 & 0.70 \\ 
\hline
Single $p$-value Method with Layer 7 & 0.77 & 0.83 & 0.93 & 0.94 \\
Single $p$-value Method with Layer 15  & 0.61 & 0.79 & 0.78 & 0.78 \\
Single $p$-value Method with Layer 22  & 0.56 & 0.68 & 0.74 & 0.73 \\ 
\hline
Ensemble Approach with Majority Voting & 0.75 & 0.81 & 0.55 & 0.55 \\
\textbf{Ours with K=3 (Layers 7, 15, and 22)} & \textbf{0.83} & \textbf{0.91} & \textbf{0.95} & \textbf{0.96}\\
\hline
\end{tabular}
}
\vspace{-1mm}
\caption{\label{tab:main-results}
AUROC ($\uparrow$) results for EYE-LLaMA and MentaLLaMA on both OOD datasets. Best results are in bold.}
\end{table*}

% \vspace{-5mm}

% \vspace{-3mm}

% \vspace{-3mm}

% We also plot ROC curves in \Cref{fig:main-plots} for the single $p$-value method and our approach. These are also averaged over $5$ runs of random iD test and calibration datasets.  
% The ROC curves are averaged over multiple folds, using a randomly sampled subset of the in-domain data for calibration in each fold. The method is evaluated on a mix of the remaining in-domain queries and OOD data. 
% Results are averaged across folds, with AUROC showing minimal variance—usually as low as 0.001.
% Lastly, we plot the false alarm curves in \Cref{fig:guarantee-curves}, illustrating the false positive rate at different thresholds of confidence. Our method is able to provide reasonable false alarm guarantees compared to the confidence threshold $\epsilon$.

%\input{tables/main-results}

%\\
\subsection{Ablation Studies} 

We perform the following studies relevant to the proposed Algorithm~\ref{alg:ood_detection}.

\paragraph{1. Choice of the Valid Merging Function:} As mentioned in \Cref{sec:combining-hypothesis}, different valid merging functions can be used to combine the $K$ $p$-values. We use Arithmetic Mean on $p$-values from the three layers ($7$, $15$, and $22$) for reporting the results in Section~\ref{main_results}. We also experiment with the other three functions and report the AUROC  results in \Cref{tab:agg-p-value}. We observe that Arithmetic Mean performs the best with comparable performance by Geometric Mean. Performance by all the functions is comparable in the case of MentaLLaMA. ROC curves for the other three valid merging functions (with similar performance as Arithmetic Mean) are included in the Appendix. 

\input{tables/aggregate_p_values}

% Ablation - different layer combinations. 
% even single layers can be ablation, as well as baseline.
% Ablation - different p-value combination methods than mean.

\textbf{2. Unchanged Responses:}
%\textcolor{blue}{RK: would 'unchanged' be a better word than 'broken'? If so, will have to change it in the intro dia as well.}
We set an upper limit of $m = 30$ on the maximum number of neurons to be dropped in Algorithm~\ref{alg:ood_detection}. It is possible that the response to a particular query might not be changed even after dropping all the $m$ maximally activated neurons. For example, while running dropout on layer 7 of EYE-LLaMA on the EyeQA calibration set, we observe that approximately $91\%$ of responses changed with the number of dropped neurons $i \leq m$. However, this percentage varies by layer: $74\%$ for layer $15$ and $56\%$ for layer $22$. 
% This trend of dropout triggering the flipping of a response more in the earlier layers compared to later layers is observed throughout our experiments. Our hypothesis is that the neurons in the earlier layers are essential to understand the meaning of the query, also hinted at by \citet{stages_of_inference}, resulting in the model being more sensitive to dropout in earlier layers.
Across our experiments, dropout in earlier layers more often changes responses than in the later layers.
We hypothesize this is because early layers are crucial for understanding the query, as also suggested by \citet{stages_of_inference}, resulting in the model being more sensitive to dropout in earlier layers.

For the proposed algorithm, the non-conformity score, and hence the $p$-value is undefined if a response is not changed within this upper limit of $m$. When aggregating the $p$-values from multiple layers, we only consider those layers where the $p$-value is well defined. If no $p$-value is defined for any of the layers, we resort to a default prediction of in-distribution for that particular query. The idea being if the response does not change even after dropping all the $m$ maximally activated neurons in all the $K$ layers, then the model is highly robust to the input query: a property more likely to be for iD queries than OOD. This default prediction, however, occurs very rarely: only $3\%$ of the total queries for Eye-LLaMA. Due to the ensemble approach, our method is able to make a prediction through one of the layers most of the time.

\textbf{3. Ablation on $\textbf{m}$:} 
In our experiments, we iteratively drop up to a maximum of $m=30$ neurons. We analyze the effect of varying this limit on a subset of the EYEQA dataset. Specifically, we fix layer $15$ (middle layer) as the dropout layer, and run Algorithm~\ref{alg:ood_detection} on this subset with $m=10, 30,$ and $50$. We observe that responses for $59\%, 78\%$, and $81\%$ of total queries are changed within these limits, respectively. As expected, we observe that a higher $m$ is more likely to change a response. It should be noted that increasing $m$ increases the number of iterations in the algorithm. We choose $m=30$, offering a balance between the fraction of changed responses and computational efficiency.

% This observation leads us to hypothesize that there are some `easier to change' queries, which change at lower limits, and some `harder to change' queries which only change at higher limits. 

\textbf{4. Difficulty Level of Queries:}
Based on the number of dropped neurons required to change the response, we also try to categorize queries. 
A venn diagram categorizing queries in the sets based on when they change is shown in \Cref{fig:diff-limits-venn} of the Appendix. 
%Responses changing in a lower value of $m$ also change within the higher $m$, as expected. However, we
We analyze two extreme `sets' of queries: `Set A' with queries whose responses were changed at all $m=10, 30, \ \text{and}\  50$ ($500$ queries) and `Set B' with queries whose responses changed only at $m=50$ ($54$ queries).
We observe that in Set A, roughly $41$\% of the queries are MCQs, requiring a choice from given options. On the other hand, Set B has only $10\%$ of such choice-based questions. 
We further check the proportion of choice-based questions in `Set C' - containing responses altered at $m=30$ and $50$ but not at $m=10$ - which is $17\%$: in between that of Sets A and B. 
% A visualization of these categories is shown in \Cref{fig:diff-limits-venn} in the Appendix. 
%\textcolor{blue}{RK: Should we say what is in Set C?}
%\AG{Set C is the 22.5\% set}

Thus, we observe that MCQs are more easily altered, requiring fewer dropped neurons compared to subjective queries. This is an interesting observation, which also seems to be supported by our results in \Cref{tab:main-results}, where we perform better when the OOD dataset is MedMCQA: a set containing entirely MCQs, as compared to COVID-QA, containing entirely subjective queries.

% \begin{figure*}[!t]
%     \centering
%     \includegraphics[width=0.8\linewidth]{images/categories.png}
%     \caption{
%     Ablation on the difficulty level of queries measured by varying the maximum number $m$ of dropped neurons in the Layer $15$ of EyeLLaMA model. 
%     We observe that responses to MCQs are more easily altered, requiring fewer neurons to be dropped compared to subjective queries. As the query becomes more subjective, it requires more neurons to change the response.
%     %\AG{Label the sets A, B and C to easily refer in text.}
%     }
%     \label{fig:diff-limits-venn}
%     % \vspace{-0.5cm}
% \end{figure*}

%% file: tables/aggregate_p_values.tex
% \usepackage{graphicx}

% \begin{table}
% \centering
% \resizebox{0.6\linewidth}{!}{%
% \begin{tabular}{c|c}
% Aggregation method & \multicolumn{1}{l}{AUROC} \\ 
% \hline
% Bonferroni Method & 0.67 \\
% Harmonic mean & 0.76 \\
% Geometric mean & 0.8 \\
% Arithmetic Mean & 0.83
% \end{tabular}
% }
% \caption{\label{tab:agg-p-value}
% AUROC for different aggregation methods. These results are on the EYE-LLaMA model with CovidQA as the OOD dataset.
% \AG{Add other dataset/model combinations?}
% \AG{Add plots for other aggregation methods in appendix}
% }
% \end{table}

% 
\begin{table}
\centering
\resizebox{\linewidth}{!}{%
\begin{tblr}{
  cells = {c},
  cell{1}{2} = {c=2}{},
  cell{1}{4} = {c=2}{},
  vline{2,4} = {1}{},
  vline{2,4} = {2-6}{},
  hline{3} = {-}{},
}
\hline
\textbf{Model}            & \textbf{EYE-LLaMA} &                  & \textbf{MentaLLaMA} &    \\
\hline
\textbf{OOD Dataset}          & \textbf{CovidQA}   & \textbf{MedMCQA} & \textbf{CovidQA}     & \textbf{MedMCQA} \\
\textbf{Bonferroni Method} & 0.67               & 0.79             & 0.93                 & 0.93             \\
\textbf{Harmonic Mean}     & 0.76               & 0.85             & 0.94                 & 0.94             \\
\textbf{Geometric Mean}    & 0.80                & 0.89             & \textbf{0.95}                 & 0.95             \\
\textbf{Arithmetic Mean}   & \textbf{0.83}               & \textbf{0.91}             & \textbf{0.95}                 & \textbf{0.96} \\
\hline
\end{tblr}
}
\vspace{-0.25cm}
\caption{\label{tab:agg-p-value}
AUROC by the proposed approach with different valid $p$-value merging functions.
}
\vspace{-0.5cm}
\end{table}

%% file: 2_rel.tex
\section{Related Work}
\label{sec:rel}
OOD detection has been of significant research focus for reliable deployment of traditional deep learning as standalone models~\citep{baseline, mahalanobis, csi, kaur, unsuper_2, semanticOODdet} or as learning enabled components in closed-loop systems~\citep{cai2020real, yang2022interpretable, ramakrishna2022efficient, beta-vae-2, yang2024memory}. Some of these approaches are built on ICAD for providing false alarm guarantees~\citep{idecode, cai2020real, codit, kaur2024tcps, yang2024memory}. MC-dropout has also been used as the Bayesian inference approach for quantifying uncertainty in traditional deep learning models~\citep{gal2016dropout, ryu2019uncertainty}.

For LLMs, main focus has been on quantifying uncertainty in the models' responses~\citep{self_prob, lin, kuhn, uq_survey, kaur2024addressing, padhi2025calibrating}. Conformal prediction has been also used to generate sets instead of a single prediction to account for uncertainty in LLM's predictions~\citep{ conformal_lang_modeling, wang2025copu}. 

OOD detection for LLMs has started to emerge for specialized LLMs such as conditional language models~\citep{ood_conditional_llms}, models fine-tuned with LoRA adapters~\citep{lora_ood_det}, and for specific tasks like sentiment analysis~\citep{sentiment_ood_det}. 
% In contrast, our approach is model and application-agnostic, making it applicable to any LLM specialized for a particular domain. Also, unlike existing approach by~\citet{sentiment_ood_det}, the proposed OOD detection algorithm does not require any learning and can be directly applied at inference time. To the best of our knowledge, this work is the first one on OOD detection for LLMs with theoretical guarantees.
In contrast, our method is model- and application-agnostic, applicable to any domain-specialized LLM. Unlike existing works like ~\citet{sentiment_ood_det}, it requires no learning and operates directly at inference. To the best of our our knowledge, this is the first OOD detection method for LLMs with theoretical guarantees.

% These approaches use the difference in the statistical or geometric properties of in-distribution and OOD data for detection. In case of LLMs, 

% In case of LLMs, training distribution is not well defined..so people have considered specific scenarios - fine-tuned models/particular application/CLMs.
% \begin{itemize}
%     \item OOD detection in LLMs: \url{https://openreview.net/pdf?id=H7Q5hHcvZE}: use the distribution in the LoRA embeddings to detect OODs - specific to LoRA fine-tuned model.
% \url{https://www.sciencedirect.com/science/article/abs/pii/S0950705125000231}: detecting textual OOD data in sentiment analysis using large language models (LLMs) - specific to sentiment analysis in text domain, \url{https://arxiv.org/pdf/2209.15558}: We present
% a highly accurate and lightweight OOD detection method for Conditional LMs, and demonstrate its effectiveness on abstractive summarization and translation

    % \item Dropout was mainly used for avoiding overfitting: \url{https://www.jmlr.org/papers/volume15/srivastava14a/srivastava14a.pdf}, \url{https://www.mdpi.com/2079-9292/12/14/3106}, dropout for OOD detection: \url{https://epubs.siam.org/doi/pdf/10.1137/1.9781611978520.63} - Two
% traditional OOD detection methods, MC-Dropout (Dropout)[10][27].
    % \item Polysemanticity: SAE paper - \url{https://arxiv.org/abs/2309.08600}. Then droplout and polysemantic neurons: \url{https://arxiv.org/pdf/2401.17975}.
% \end{itemize}

%% file: 6_conc.tex
\section{Conclusion}
\label{sec:conc}
In this paper, we introduce a novel, model-agnostic conformal OOD detection method for specialized LLMs, leveraging dropout tolerance as a non-conformity measure within the ICAD framework. Our inference-time OOD detection approach aggregates dropout tolerance across multiple layers using valid ensemble merging functions, preserving theoretical false alarm guarantees. Extensive experiments with medical-specialized LLMs demonstrate that our method consistently outperforms baseline approaches—achieving AUROC gains of up to 37\%—and adapts to a variety of OOD dataset types and query complexities. Furthermore, our ablation studies provide additional insights into the effects of various design choices, informing the method’s applicability, generalizability, and practical utility. Moving forward, this work opens avenues for extending conformal OOD detection to multi-modal LLMs and broader deployment in safety-critical applications.

%% file: appendix_tex.tex
\section{Appendix}
\label{sec:appendix}

\begin{figure*}[!t]
    \centering
    \includegraphics[width=1\linewidth]{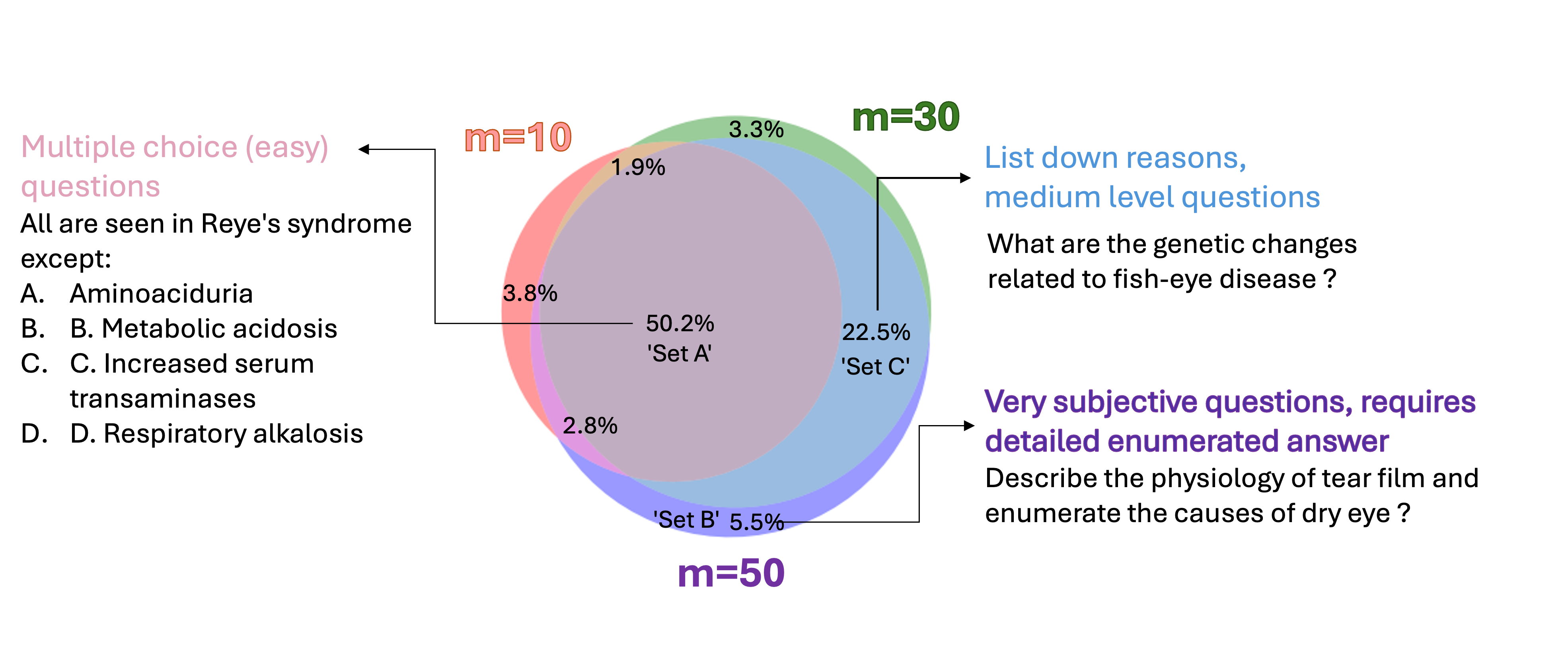}
    \caption{
    Ablation on the difficulty level of queries measured by varying the maximum number $m$ of dropped neurons in the Layer $15$ of EyeLLaMA model. 
    We observe that responses to MCQs are more easily altered, requiring fewer neurons to be dropped compared to subjective queries. As the query becomes more subjective, it requires more neurons to change the response.
    %\AG{Label the sets A, B and C to easily refer in text.}
    }
    \label{fig:diff-limits-venn}
    % \vspace{-0.5cm}
\end{figure*}

\subsection{Implementation details}
% \AG{Complete - maybe in appendix?}
% Exact number of neurons, step size, layers used, calib-test split (ratio), 
%\AG{Stages of inference paper - how you decide the layers to drop.}
We set the limit on the maximum number of neurons to be dropped, $m = 30$ in our main experiments. We set the step size $n = 5$, the incremental number of neurons dropped in each successive iteration of dropout.
We used GPT-4o-mini to evaluate whether the pre-dropout response is semantically similar to the post-dropout response.
Inspired by~\cite{stages_of_inference}, we choose layers $7$, $15$, and $22$ to perform the dropout - each of these layers lies in a different stage of inference, and hence the model has a different understanding of the query at each of these individual layers. 
We use a random $20\%-80\%$ calibration-test split on the in-domain data for each run.  We load the LLMs in $8$-bit precision and perform inference on A100 GPUs. We use PyTorch hooks to get internal activations of the LLMs and trigger dropout. 

Also, both EYE-LLaMA and MentaLLaMA are open-source and publicly available to be used for research purposes under MIT license.

\subsection{More evaluation results}
We have already evaluated with CovidQA and MedMCQA as the OOD datasets. However, the EYE-QA and IMHI dataset are also OOD for MentaLLaMA and EYE-LLaMA respectively. In this section, we evaluate the AUROC for these model-dataset combinations as well.

\begin{figure*}[h]
    \centering
    \includegraphics[width=0.7\linewidth]{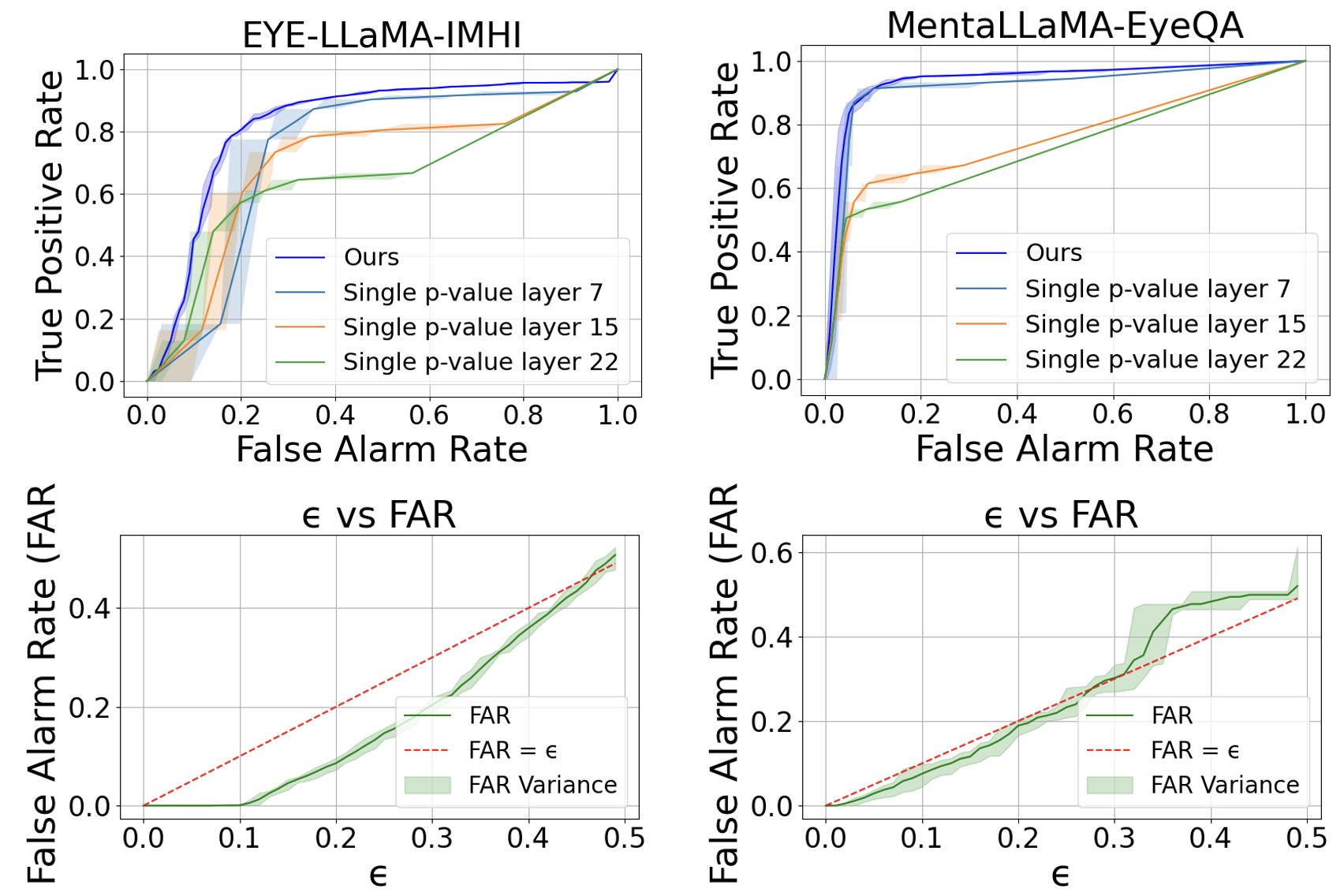}
    \caption{
    \label{fig:id_plots_roc}
    ROC and False Alarm  Guarantee curves on more datasets, using Arithmetic Mean as the valid merging function.
    }
    
\end{figure*}

The ROC and false alarm curves for these combinations are shown in \Cref{fig:id_plots_roc}.
Our method achieves an AUROC of 0.93 for the MentaLLaMA - EyeQA combination, and 0.82 for EYE-LLaMA - IMHI combination. This shows that our method is not specific to particular OOD datasets - even the datasets that are in-domain for one model can be used as OOD for another model, and our method can still detect them as OOD.

\subsection{Analysis of different merging functions}
In \Cref{tab:agg-p-value}, we provide the average AUROC values for when we use different merging functions in our method. 
Here, we plot the ROC and False alarm curves for each of these valid merging functions.

\begin{figure*}[!t]
    \centering
    \includegraphics[width=\linewidth]{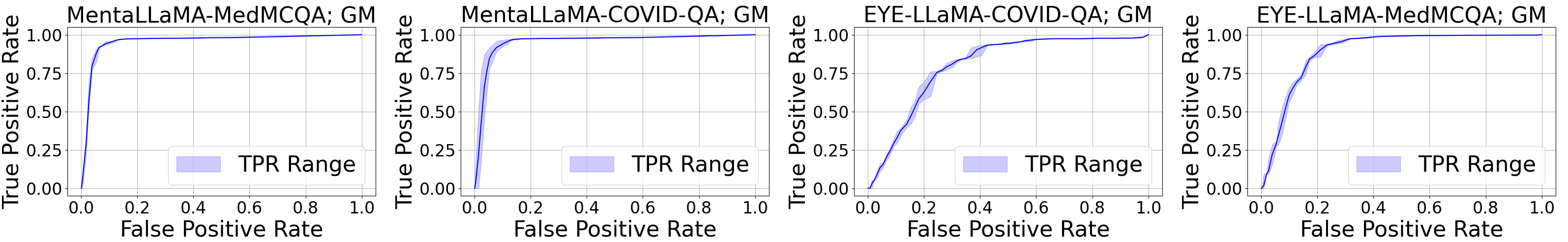}
    \caption{ROC curves with Geometric Mean as the valid merging function.}
    \label{fig:gm}
\end{figure*}

\begin{figure*}[!t]
    \centering
    \includegraphics[width=\linewidth]{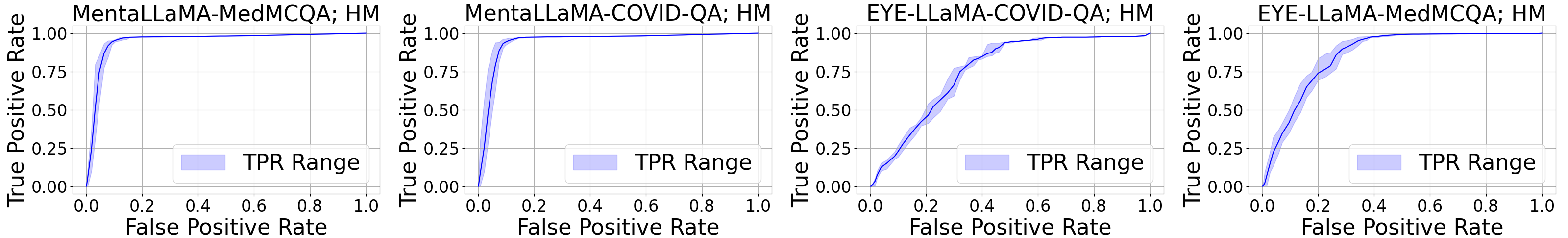}
    \caption{ROC curves with Harmonic Mean as the valid merging function.
    %\AG{Remove false alarm curves}
    }
    \label{fig:hm}
\end{figure*}

\begin{figure*}[!t]
    \centering
    \includegraphics[width=\linewidth]{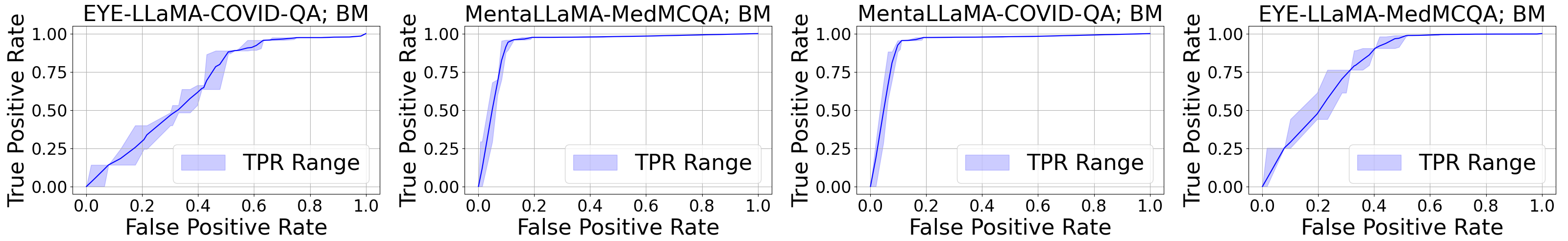}
    \caption{ROC curves with Bonferroni Method as the valid merging function.}
    \label{fig:bonferroni}
\end{figure*}

\Cref{fig:gm} shows the plots for Geometric mean, \Cref{fig:hm} for Harmonic mean and \Cref{fig:bonferroni} for the Bonferroni merging function.
We note that while there is not a significant difference in the AUROC regardless of the merging function used, the false alarm curves have a distinct shapes depending on the merging function. Specifically, the Bonferroni method is seen to have a larger false alarm rate for a given $\epsilon$ than other methods, potentially because it is more sensitive to extremely low p-values than other methods.

